\documentclass{article}

     \PassOptionsToPackage{numbers, compress}{natbib}

     \usepackage[final]{neurips_2019}

\usepackage[utf8]{inputenc} 
\usepackage[T1]{fontenc}    
\usepackage{hyperref}       
\usepackage{url}            
\usepackage{booktabs}       
\usepackage{amsfonts}       
\usepackage{nicefrac}       
\usepackage{microtype}      
\usepackage{caption}
\usepackage{natbib}
\usepackage{graphicx}
\usepackage{subcaption}
\title{Extending the Forward Forward Algorithm}

\author{%
  Saumya Gandhi\\
  \texttt{ssgandhi} \\
   \And
   Ritu Gala \\
   \texttt{rgala} \\
   \AND
   Jonah Kornberg \\
   \texttt{jkornber} \\
   \And
   Advaith Sridhar \\
   \texttt{advaiths} \\
}

\begin{document}

\maketitle

\begin{abstract}
The Forward Forward algorithm, proposed by Geoffrey Hinton in November 2022, is a novel method for training neural networks as an alternative to backpropagation.  In this project, we replicate Hinton's experiments on the MNIST dataset, and subsequently extend the scope of the method with two significant contributions. First, we establish a baseline performance for the Forward Forward network on the IMDb movie reviews dataset. As far as we know, our results on this sentiment analysis task marks the first instance of the algorithm's extension beyond computer vision. Second, we introduce a novel pyramidal optimization strategy for the loss threshold - a hyperparameter specific to the Forward Forward method. Our pyramidal approach shows that a good thresholding strategy causes a difference of upto 8\% in test error. \footnote{Our code can be found here: \href{https://github.com/Ads-cmu/ForwardForward}{https://github.com/Ads-cmu/ForwardForward}} Lastly, we perform visualizations of the trained parameters and derived several significant insights, such as a notably larger (10-20x) mean and variance in the weights acquired by the Forward Forward network.
\end{abstract}

\section{Introduction}
Backpropagation is the most widely used optimization algorithm for training neural networks today. However, while widely successful, the backpropagation algorithm has 3 important limitations. 

First, backpropagation is biologically implausible. There is no convincing evidence that the cortex of the brain explicitly propagates error derivatives or stores neural activities for use in a subsequent backward pass \cite{https://doi.org/10.48550/arxiv.2212.13345}. Moreover, backpropagation through time (the standard technique for training RNNs) is especially implausible, as the brain does not freeze in time in order to update neural connections. 

The second major drawback with backpropagation is the need for perfect knowledge of the forward pass computation in order to compute the correct derivatives. This prevents us from being able to insert "black boxes" or non-differentiable components in the neural network.

Lastly, the need to store forward pass computations and backpropagate errors across layers makes backpropagation power and memory intensive. In order to train really large networks without consuming much power, different methods for training networks will need to be explored.

Geoffrey Hinton proposed the Forward Forward algorithm in November 2022, with the goal of enabling neural networks to learn continuously without the need for backpropagation \cite{https://doi.org/10.48550/arxiv.2212.13345}. In his paper, Hinton suggests two significant advantages of the forward-forward algorithm over backpropagation. First, it provides a more plausible model of learning in the human brain, and second, it can make use of very low-power analog hardware, thereby enabling much larger networks to be trained with much less power.


This project investigates the performance of the Forward Forward algorithm in training neural networks. The key contributions of our work are as follows. First, we have replicate Hinton's original results on the MNIST dataset. Next, we establish a baseline performance for the Forward Forward network on the IMDb movie reviews dataset. As far as we are aware, our results on this sentiment analysis task mark the first instance of the algorithm’s extension beyond computer vision. Lastly, we propose a new hyperparameter optimization strategy for tuning the loss threshold of the Forward Forward network. This pyramidal optimization strategy yields a 11\% reduction in network error rate. 

Apart from the above, we also report two more results. First, we performed extensive ablations on various activation functions for the FF algorithm and report some negative results. Second, we perform some visualisation of the weights learnt by the FF algorithm and record our observations. A detailed discussion of these results, as well as hypotheses that explain them are as left as future work.  

\section{Literature Review}
\subsection{Other forward-pass based approaches to training neural networks}
A primary objective of the Forward Forward algorithm is to emulate learning processes observed in human brains. To achieve this, the algorithm adheres to a brain model known as Predictive Coding (PC) \cite{helmholtz1924treatise}, which characterizes the brain as a predictive system where each layer strives to enhance the accuracy of its own inputs. As each layer in the Forward Forward algorithm adjusts its gradients based on input data, the algorithm can be regarded as an application of PC in the domain of machine learning. Various forms of PC have previously been investigated in machine learning \cite{millidge2021predictive}. Specifically, supervised predictive coding greatly resembles the forward forward algorithm, as it only involves forward passes up and down the network (from the data to the labels and vice-versa). The recently developed Predictive Forward Forward Algorithm \cite{ororbia2023predictive} extends and integrates the concepts of FF and PC, resulting in a robust neural system capable of learning a representation and generative model. Preliminary results on the MNIST dataset suggest the potential of this brain-inspired, backpropagation-free approach for credit assignment within neural systems.

Another forward pass based approach was proposed by Dellaferrera et al. \cite{dellaferrera_error-driven_2022} They propose a learning rule that replaces the backward pass with a second forward pass in which the input signal is modulated based on the error of the network. This learning rule addresses various issues such as weight symmetry, dependence of learning on non-local signals and the freezing of neural activity during error propagation. The authors demonstrate the effectiveness of their approach on MNIST, CIFAR-10, and CIFAR-100 datasets.

\subsection{Directional approaches to training neural networks}
Schmidhuber et al. \cite{kirsch_meta_2022} propose the Variable Shared Meta Learning (VSML) algorithm, which unifies various meta-learning approaches. The authors demonstrate on the MNIST and CIFAR10 datasets that simple weight-sharing and sparsity in an NN can express powerful learning algorithms in a reusable fashion. They implement the backpropagation learning algorithm solely by running in forward-mode, eliminating the need for a backward pass.

Baydin et al. \cite{baydin_gradients_2022} present a method for computing gradients based solely on the directional derivative that one can compute exactly and efficiently via the forward mode. They call this formulation the forward gradient. They demonstrate forward gradient descent on the MNIST dataset, showing substantial savings in computation and enabling training up to twice as fast in some cases.

\section{Dataset Description}
Our baseline implementation and threshold ablations were performed on the MNIST dataset \cite{726791}. In order to demonstrate the extensibility of the Forward Forward network to other domains, we picked a sentiment analysis task using the IMDb reviews dataset \cite{maas-EtAl:2011:ACL-HLT2011}. The dataset contains 25,000 positive and 25,000 negative movie reviews, split equally into test and train datasets. Each review was preprocessed by removing HTML tags, stop words and by performing stemming. Next, each word of the review was passed through Word2Vec \cite{mikolov_efficient_2013} to get a lower dimensional representation of the word. Word2Vec uses a 2 layer neural network with no non-linearity in the hidden layer, and therefore can be approximated as a single layer network. Single layer neural networks do not backpropagate gradients and hence Word2Vec is an acceptable feature extractor for the Forward Forward network.  
\subsection{Generating Positive and Negative Data}
MNIST images must have labels embedded into them before they can be passed through the Forward Forward network. This is done by utilising the black border around MNIST images. In order to append label data to images, we set the pixel corresponding to the label amongst the first 10 pixels to 255, and reduce the rest to a value of 0. These images, with correctly appended labels, constitute positive data for the model. For our sentiment analysis task, the positive/negative label was one-hot concatenated to the Word2Vec feature vector. 

The Forward Forward algorithm also requires negative data during training. Negative data is generated by randomly appending the wrong label to the input image/review before passing it through the network. We use an equal number of positive and negative samples during training. 
\begin{figure*}[t!]
    \centering
    \begin{subfigure}[t]{0.5\textwidth}
        \centering
        \includegraphics[width=\textwidth]{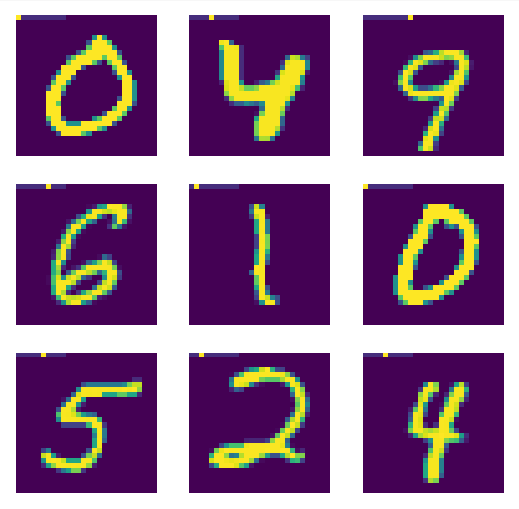}
        \caption{Positive data samples}
    \end{subfigure}%
    ~ 
    \begin{subfigure}[t]{0.5\textwidth}
        \centering
        \includegraphics[width=\textwidth]{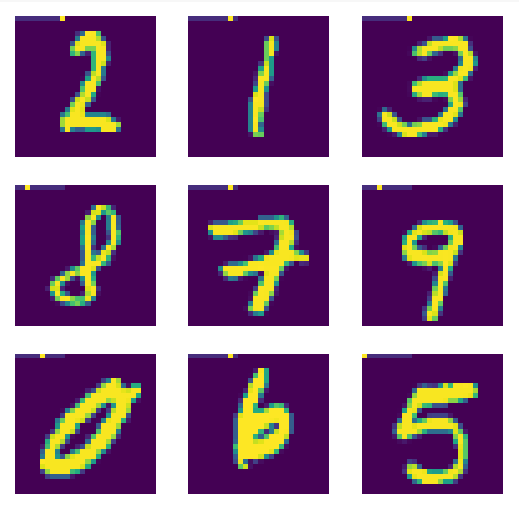}
        \caption{Negative data samples}
    \end{subfigure}
    \caption{Data Generation for the Forward Forward Network. The appended label can be seen in the first 10 pixels of the image}
\end{figure*}

\begin{figure*}[t]
    \centering
        \includegraphics[width=\textwidth]{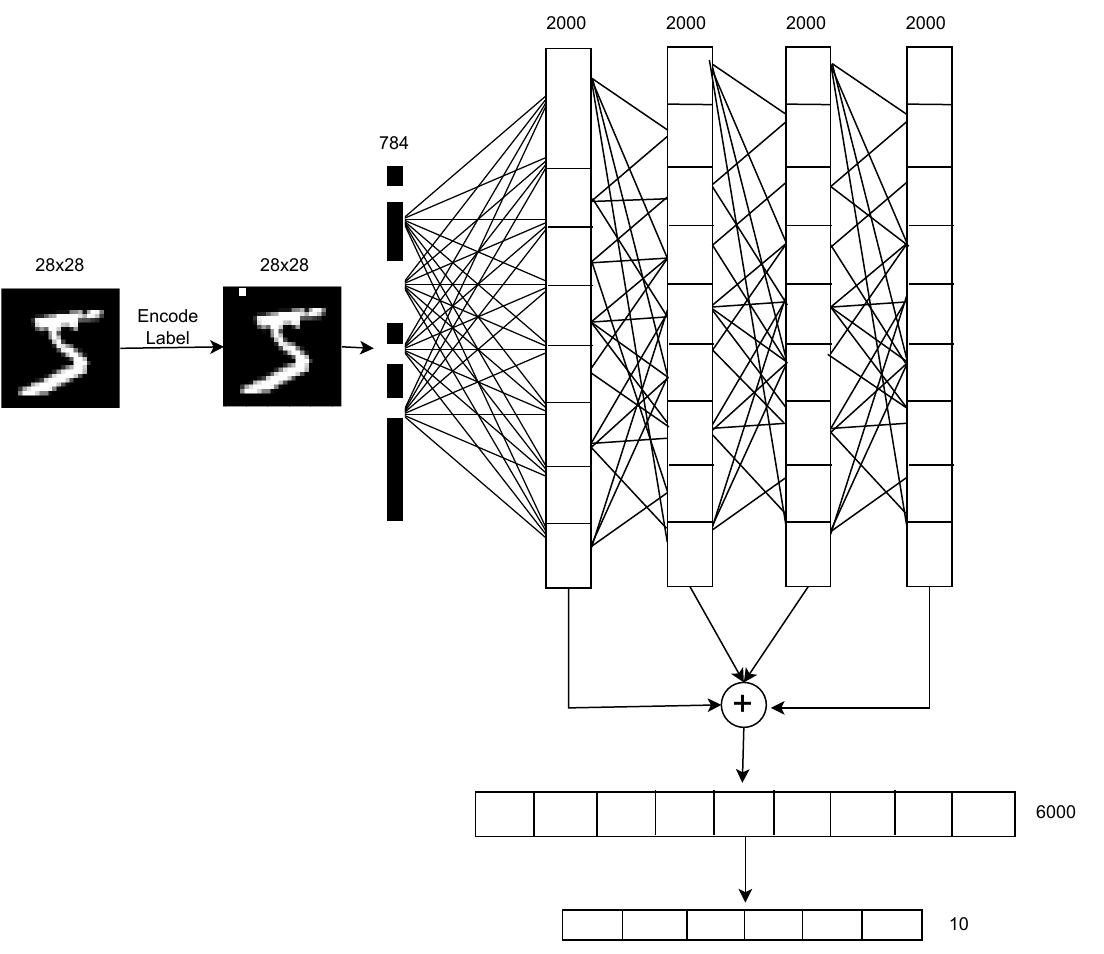}
    \caption{Forward Forward model architecture. Although this looks similar to a normal feedforward network, it is important to remember that gradients are detached from every previous layer. The additional component at the bottom is a single layer neural network that takes in concatenated layer activations as input and predicts a class label}
\end{figure*}

\section{Model Description}
\subsection{Layer Training}
Our network consists of several layers, each with its own loss function. The goal of the loss function is maximise the layer activation for positive data while minimising the layer activation for negative data. More concretly, the training loss for each layer is the difference between the sum of neuron activations to positive/negative inputs and a threshold hyper-parameter value. This threshold hyperparameter is known as the loss threshold, and we perform extensive hyperparameter tuning of this threshold in our analysis. It is worth nothing that during the forward pass, the input is normalized to prevent its magnitude from affecting the layer's output activation magnitude. 

\subsection{Network Architecture}
As shown in Figure 2, the network consists of 4 fully connected layers with 2000 neurons each. Note that there is no output label layer. We use the Adam optimizer to optimize the network.

\subsection{Inference}
There are 2 plausible methods of inference for the Forward Forward network. The first method involves using a 1 layer classification neural network that uses the FF network activations as features for its classification task. An alternate method involves appending a label to the input image and passing it through the network. This process is repeated 10 times (once for each label), and the label that produces the maximum activation is chosen as the output label for the image. Between the two methods, we report results for the approach that uses the 1 layer classification network as it emperically provides better results. 

\subsection{Baselines}
The original paper uses a fully connected neural network trained using backpropagation as a baseline network. This backpropagation-trained network has a 1.4\% test error. We were able to reproduce this baseline and achieve a similar error rate.

The original paper proposed two major ways in which networks could be trained using Forward Forward (FF): an Unsupervised and a Supervised example of FF. For this report, our focus was to reproduce the training pipeline and architecture for the Supervised example of FF. The original paper was able to get around 1.36\% with their forward forward architecture. Although the paper made it clear that the architecture they used included 4 fully connected layers with 2000 neurons each with ReLU activation, their loss function, optimizer, learning rate, threshold, and scheduling strategy was not elaborated on. Since the forward-forward learning dynamics are highly different from backpropagation, our standard intuitions and starting points did not work well. High thresholds performed better than lower ones, potentially since higher thresholds allow a wider range of squared activations for negative samples. Increasing threshold from 0.5 to 10 improved our model performance by approximately 8\%. However, this also had the side effect of significantly slowing down convergence. We hypothesize that this happened because the model's weights were unable to change fast enough to adjust to the large threshold using our low learning rate. We used the Adam optimizer initialized with a learning rate of 0.01 (unusually high for Adam) to have our model converge within 100 epochs with the high threshold of 10. Using this approach, we were able to achieve a test error of 1.37\% (comparable to backprop baselines).



\section{Results and Discussion}

\begin{table}[h]
\centering
\begin{tabular}{|c|c|c|}
\hline
\textbf{} & \textbf{MLP Backprop} & \textbf{Forward Forward} \\
\hline
\textbf{MNIST} & 1.16\%(20 epochs) & 1.37\%(63 epochs) \\
\hline
\textbf{Movie Reviews} & 15.3\% (3 epochs) & 15.14\% (6 epochs) \\
\hline
\end{tabular}
\label{tab:my_table}
\vspace{8px}
\caption{Error rates of Backprop and Forward Forward trained models on various datasets.}
\end{table}

\subsection{Forward Forward on Sentiment Analysis}
A key requirement for biological plausibility is the ability for a training algorithm to work across multiple domains such as vision and language. In this study, we investigated the performance of the Forward Forward algorithm on the IMDB movie reviews dataset, with the aim of assessing its ability to generalize beyond Computer Vision and work effectively on Natural Language Processing tasks. Our results show that the Forward Forward algorithm achieved an accuracy of 84.86\% on the test set after 6 epochs, indicating its potential in learning patterns beyond visual data. To compare its performance with traditional backpropagation-based networks, we trained a fully connected network with the same architecture as the Forward Forward network, and an output layer. We observed that the backpropagation-based network was also able to achieve an accuracy of 85\% on the test set, albeit in fewer epochs, consistent with convergence findings in computer vision tasks.

Our findings suggest that the Forward Forward algorithm can be an effective alternative to backpropagation-based networks in the context of NLP tasks. Further research is warranted to explore the performance of the Forward Forward algorithm in training embeddings from scratch, and in performing more complex NLP tasks such as language modelling.

\subsection{Threshold Ablations and Analysis on Forward Forward}

The Forward Forward algorithm introduces a new training hyperparameter - the loss threshold. Finding the appropriate threshold for this hyperparameter is crucial for the algorithm to work well. At each layer of our algorithm, the sum of squared activations for every example in our batch is calculated and compared against this loss threshold. The layer's goal is to maximize this sum so that it exceeds the threshold for positive examples and falls below the threshold for negative examples. This is important because the subtracted value is passed into our loss function, which penalizes larger values. 

Hinton, in his experiments, uses a threshold equal to the number of neurons in the layer. This can be rewritten as a threshold proportional to the number of neurons in a given layer, with a proportionality factor (k) of 1. We set this as our baseline for further study.

\begin{figure*}[h]
\centering
\includegraphics[width=\textwidth]{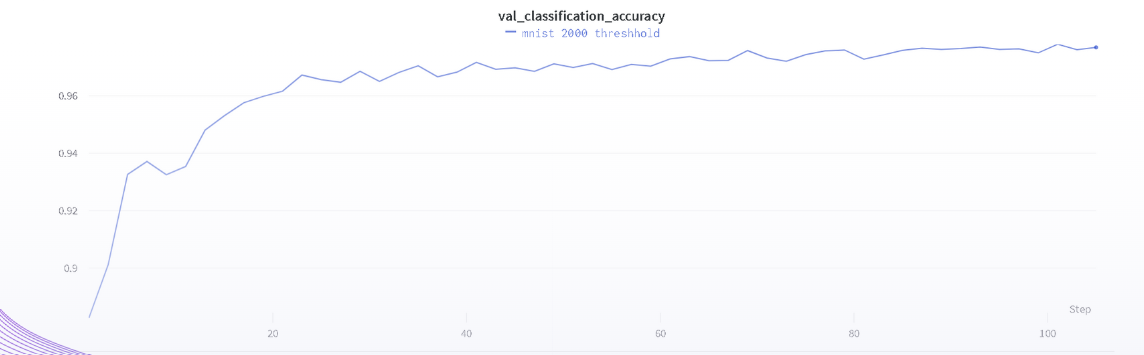}
\end{figure*}

In our initial experiments, we varied the value of k and tested values ranging from 0.005 to 10 to determine the values that give us the best results. We found that a k between 0.3 and 0.5 gives better results than our initial baseline.
\begin{figure*}[h]
\centering
\includegraphics[width=\textwidth]{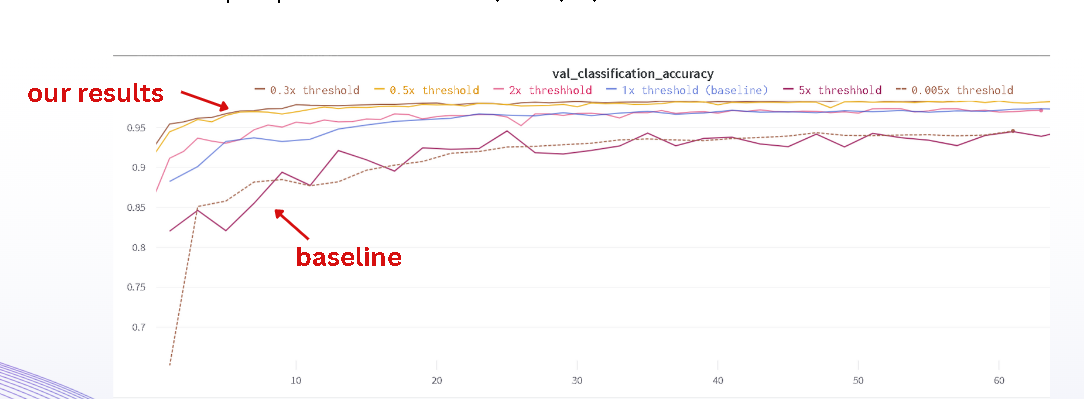}
\end{figure*}

These initial experiments led us to consider whether it would make sense to have different thresholds for different layers. We tried using monotonically increasing values of k across layers and also tested monotonically decreasing values. We found that the monotonically increasing the threshold across layers performs distinctly better than other approaches. We hypothesize that larger threshold values in later layers improves performance as the later layers are responsible for higher level feature recognition while the lower layers tend to behave as feature extractors. We refer to this monotonically increasing threshold strategy as the pyramidal approach to loss threshold tuning.
\begin{figure*}[h]
\centering
\includegraphics[width=\textwidth]{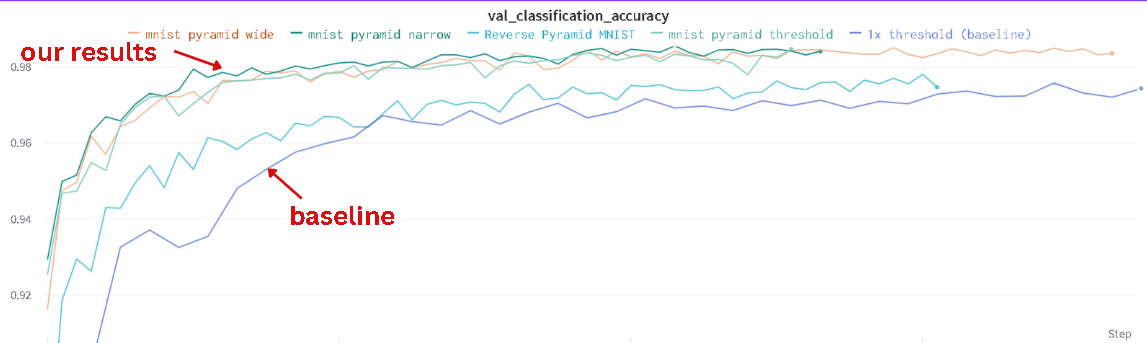}
\end{figure*}

We also explored the use of a threshold scheduler in our ablations. The intuition behind this approach is that the model might benefit from having lower thresholds initially while it is still learning, and then gradually increasing the penalty as it trains more. Our results show that using a threshold scheduler provides a promising direction compared to our baseline.
Overall we see an error reduction from 1.8\% to 1.3\% using this approach.
\begin{figure*}[h]
\centering
\includegraphics[width=\textwidth]{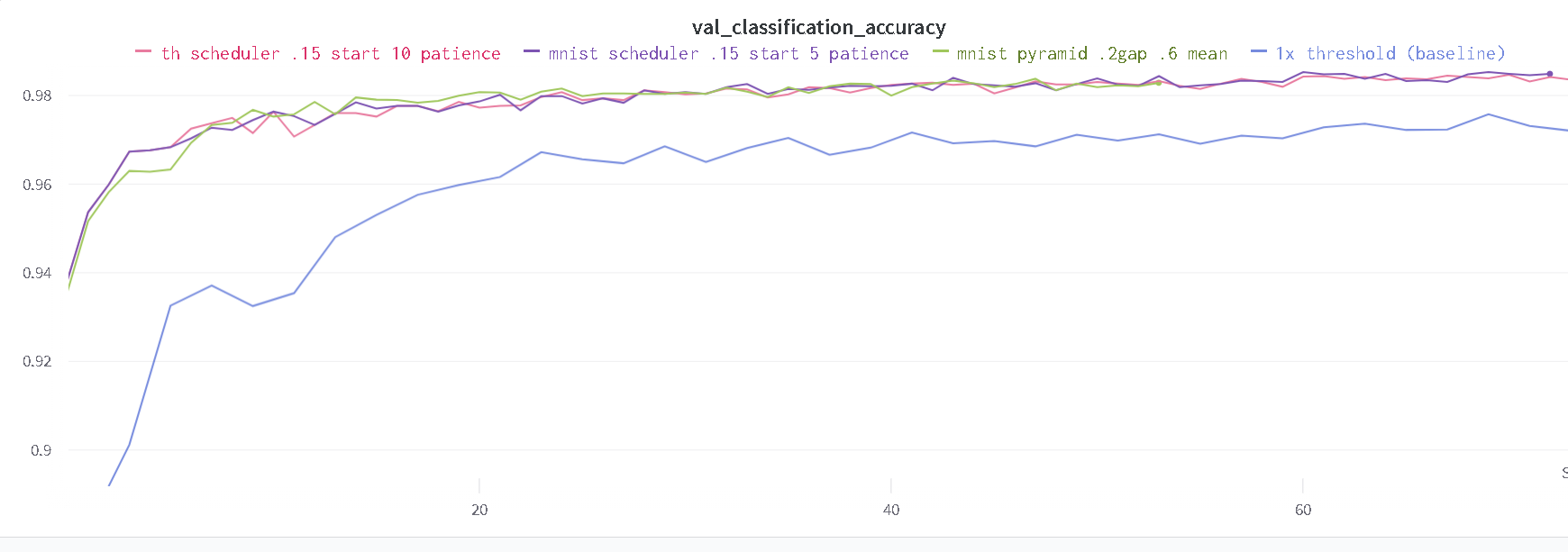}
\end{figure*}

\subsection{Activation Function Analysis}
In his forward-forward algorithm, Hinton employs the ReLU activation function. We conducted an investigation into the performance of other commonly used activation functions within the context of the forward-forward algorithm. As illustrated in the graph below, most activation functions perform well with this algorithm. However, one notable observation is that bounded activations such as tanh and sigmoid do not train at all for certain thresholds. \\
\begin{figure*}[h]
\centering
\includegraphics[width=\textwidth]{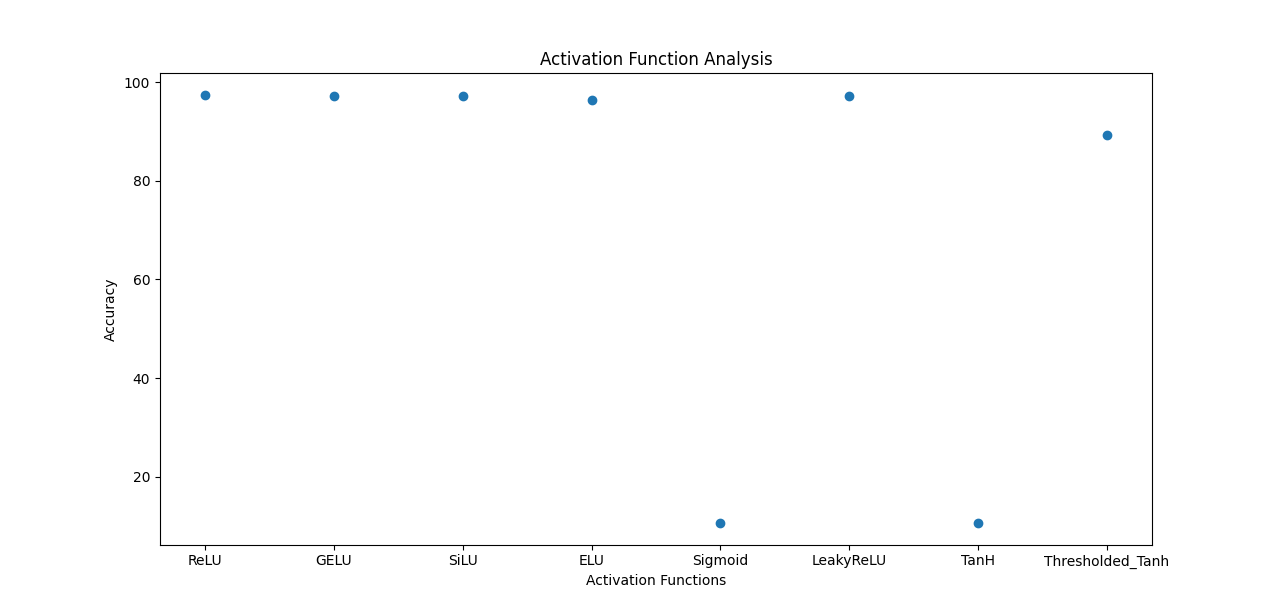}
\end{figure*}

Even after tuning the threshold hyperparameter, these activations do not perform as well as others. One possible hypothesis here may be that this is due to the nature of the objective functions - maximising a bounded activation may require incredibly high weight values, even to exceed low thresholds.
\begin{figure*}[h]
\centering
\includegraphics[width=\textwidth]{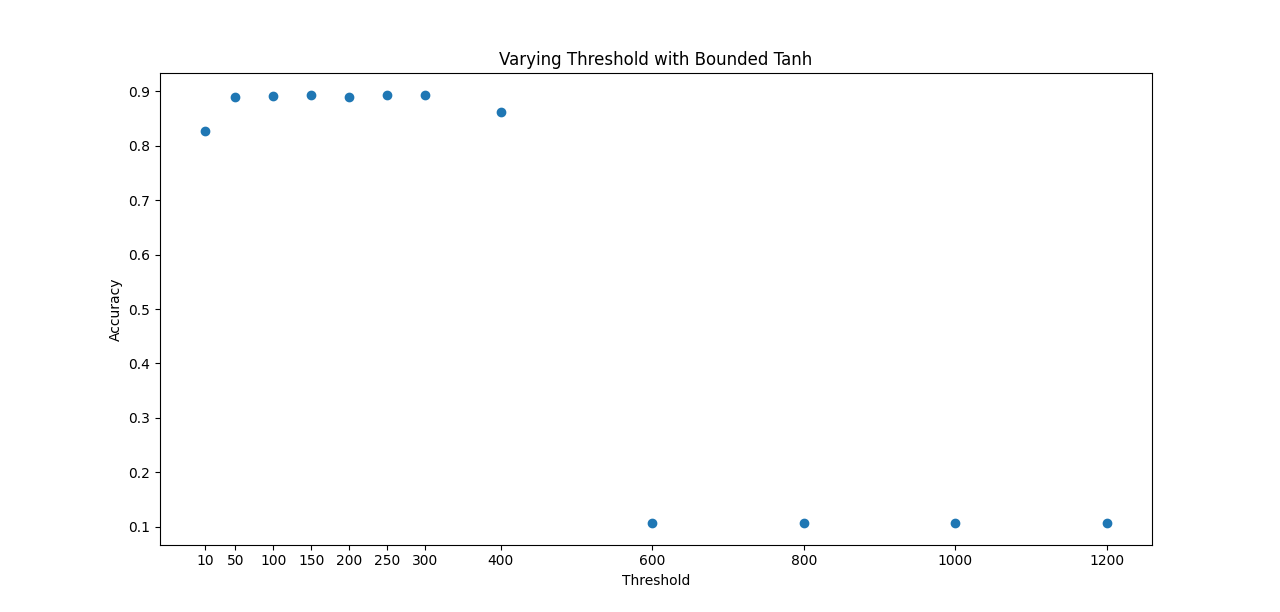}
\end{figure*}
\subsection{Weight analysis}
Lastly, we analyze the weight matrices of the trained network. Our analysis revealed several key findings. Firstly, we observe that the range of weights for the Forward Forward trained network are much larger (10-20x) than that of the backpropagation trained network, with a range of -14.36 to 18.19 compared to -0.67 to 0.43, respectively. Notably, weight decay was not applied in either case. This disparity in weight ranges may be attributed to the objective function of the Forward Forward algorithm, which aims to improve performance by encouraging highly positive activations for positive samples and highly negative activations for negative samples.

Secondly, we found that the range of weights decreased as we went deeper into the network, although the underlying reasons for this pattern requires further investigation. Finally, we observed a strong spike in the weights connected to the encoded label part of the input, which is consistent with the notion that this aspect of the input contains crucial information for the network to predict the correct label.

Taken together, our weight matrix analysis provides additional insights into the workings of the Forward Forward algorithm and suggests potential avenues for future research in understanding the mechanisms underlying its effectiveness.

\begin{table}[h]
\centering
\begin{tabular}{|c|c|c|}
\hline
\textbf{Layer number} & \textbf{Min Weight} & \textbf{Max Weight} \\
\hline
\textbf{1} & -14.36 & 18.19 \\
\hline
\textbf{2} & -11.28 & 4.95\\
\hline
\textbf{3} & -9.85 & 2.69 \\
\hline
\textbf{4} & -6.47 & 1.53 \\
\hline
\end{tabular}
\label{tab:my_table_2}
\caption{Min and max weight values for different layers of networks trained using backprop and FF algorithms}
\label{tab:my_table_2}
\end{table}

\begin{figure*}[h]
\includegraphics[width=\textwidth]{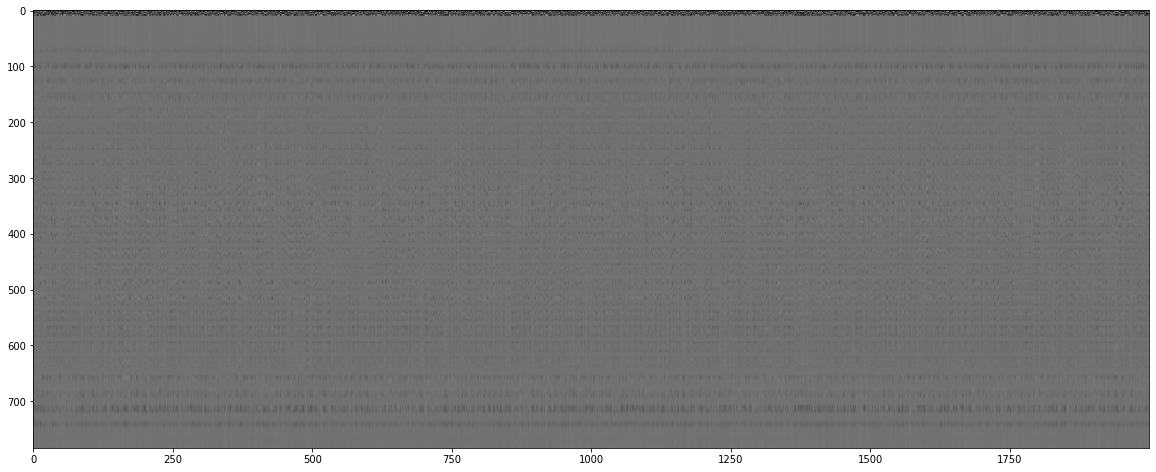}
\caption{
First hidden layer weight matrix. The first few rows represent the weights connected to the pixels that had encoded labels. Observe the sharp difference in their values as opposed to other weight values in the image.
}
\label{fig1}
\end{figure*}

\section{Conclusion}
Our study explored the effectiveness of the Forward Forward algorithm on data beyond Computer Vision and conducted experiments to understand the effects of various parameters of the algorithm. We found that the Forward Forward algorithm performs comparably to its backpropagation variant and uncovered an important relationship between the threshold parameter and the depth and size of each layer of the forward forward network. Both of these contributions are novel and warrant further interest in exploring the Forward Forward algorithm.

Our study sets the stage for further exploration of more architectures and hyptheses regarding Forward Forward. This could include more complex NLP tasks where text embeddings can be trained from scratch, and the Forward Forward algorithm can be applied over time. Additionally, future work could investigate the use of more biologically inspired activations, such as the negative log of the student T distribution that bring the algorithm even closer to biological alignment. Overall, our findings suggest that the Forward Forward algorithm holds promise as a viable alternative to backpropagation and merits further exploration in the context of various machine learning algorithms that are biologically aligned.

\bibliography{sample}

\end{document}